%% file: BMF-BID-AlphaDiversification.tex
\documentclass{article} 
\usepackage{iclr,times}

\input{symbols.tex}
\usepackage{flushend} 
\usepackage{balance}
 
 \usepackage{amsmath,amsfonts,bm}
\usepackage{array,multirow,graphicx}
\usepackage{changes}
\usepackage{booktabs}
\usepackage{graphicx} 
\usepackage{xparse}
\usepackage{hyperref}
\usepackage{url}
\usepackage{subfigure} 
\usepackage{cancel}

\usepackage{algorithm} 
\usepackage{algorithmic}

\usepackage{sidecap}
\sidecaptionvpos{figure}{t}
\usepackage{verbatimbox}
\usepackage[labelfont=bf]{caption} 
\usepackage{wrapfig}

\usepackage{color}   
\definecolor{winestain}{rgb}{0.5,0,0}
\definecolor{ocre}{RGB}{51,102,0} 
\definecolor{colorBlue2}{RGB}{200,207,248}
\definecolor{mydarkblue}{rgb}{0,0.08,0.45}
\usepackage{hyperref}
\hypersetup{
	colorlinks=true, 
	linktoc=all,     
	linkcolor=mydarkblue,  
	anchorcolor=blue,
	citecolor=mydarkblue,
}
\PassOptionsToPackage{numbers, compress}{natbib}
\usepackage[numbers]{natbib}

\renewenvironment{thebibliography}[1]{
	\begin{oldthebibliography}{#1}
		\setlength{\itemsep}{0em}
		\setlength{\parskip}{0.07em}
	}
	{
	\end{oldthebibliography}
}

\title{Feature Selection via the Intervened Interpolative Decomposition and its Application in Diversifying Quantitative Strategies}

\author{
Jun Lu \thanks{ Correspondence to: Jun Lu $<$jun.lu.locky@gmail.com$>$.
	} \\
\texttt{jun.lu.locky@gmail.com} \\
\And 
Joerg Osterrieder 
\thanks{This research was funded by the Swiss National Science Foundation within the project “Mathematics and Fintech - the next revolution in the digital transformation of the Finance industry” is gratefully acknowledged by the corresponding author. This research has also received funding from the European Union's Horizon 2020 research and innovation program FIN-TECH: A Financial supervision and Technology compliance training programme under the grant agreement No 825215 (Topic: ICT-35-2018, Type of action: CSA). Furthermore, this article is based upon work from the COST Action 19130 Fintech and Artificial Intelligence in Finance, supported by COST (European Cooperation in Science and Technology), www.cost.eu (Action Chair: Joerg Osterrieder).
\newline
Copyright 2022 by the author(s)/owner(s). September 30th, 2022.}\\
\texttt{joerg.osterrieder@utwente.nl}
}

\begin{document}

\maketitle

\begin{abstract}
In this paper, we propose a probabilistic model for computing an interpolative decomposition (ID) in which each column of the observed matrix has its own priority or importance, so that the end result of the decomposition finds a set of features that are representative of the entire set of features, and the selected features also have higher priority than others.
This approach is commonly used for low-rank approximation, feature selection, and extracting hidden patterns in data, where the matrix factors are latent variables associated with each data dimension. 
Gibbs sampling for Bayesian inference is applied to carry out the optimization. 
We evaluate the proposed models on real-world datasets, including ten Chinese A-share stocks, and demonstrate that the proposed Bayesian ID algorithm with intervention (IID) produces comparable reconstructive errors to existing Bayesian ID algorithms while selecting features with higher scores or priority.
\end{abstract}

\section{Introduction}

Over the course of the last several years, a significant amount of scholarly attention has been drawn to the issue of feature selection.
At a high level, feature selection can be considered as a branch of reducing data dimensionality of which the two primary methods are \textit{feature learning} and \textit{feature selection}.
The problem of feature learning involves the creation of new features from the original data. In contrast, the feature selection problem does not change the original representation of the data variables, so the physical meaning of each variable is preserved.
To be more specific, the feature selection problem can be subdivided into two scenarios:
supervised and unsupervised. Since we do not have target variables, selecting unsupervised features is more challenging. Typically, the unsupervised feature selection relies on matrix decomposition
\citep{cheng2005compression, liberty2007randomized, martinsson2011randomized, lu2022bayesian}, filter \citep{dash2002feature}, and embeddings \citep{dy2004feature, hou2011feature}.

On the other hand, matrix decomposition algorithms such as QR decomposition, and singular value decomposition have been used extensively over the years to reveal hidden structures of data matrices in scientific and engineering areas such as collaborative filtering \citep{marlin2003modeling, lim2007variational, mnih2007probabilistic, lu2022matrix, lu2022bayesian}, recommendation systems \citep{lu2022matrix}, clustering and classification \citep{li2009non, wang2013non}.
Low-rank matrix approximations are therefore essential in data science.
Due to the Eckart-Young-Misky theorem, low-rank approximation problems can be easily solved with singular value decomposition \citep{golub1987generalization}. However, it is frequently desirable for many applications to operate with a basis consisting of a subset of the original columns of the observed matrix \citep{martinsson2011randomized, kakushadze2016101}. 
The interpolative decomposition (ID) is one of these low-rank approximations; it reuses columns from the observed matrix, preserving matrix sparsity and nonnegativity while removing redundant information.

\begin{figure*}[h]
\centering  
\vspace{-0.35cm} 
\subfigtopskip=2pt 
\subfigbottomskip=9pt 
\subfigcapskip=-5pt 
\includegraphics[width=0.95\textwidth]{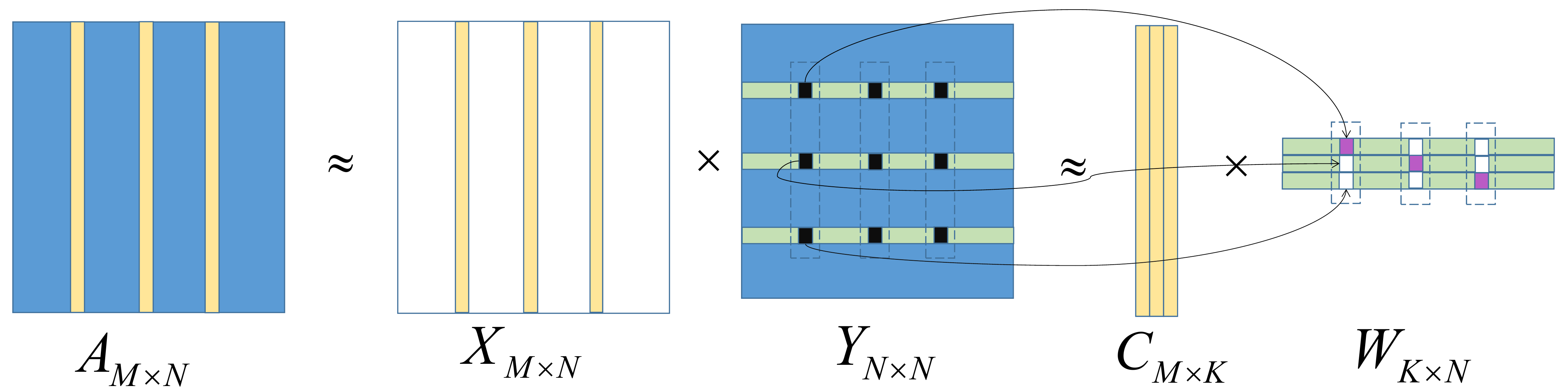}
\caption{Overview of the ID of the matrix $\bA\in\real^{M\times N}$ where the yellow vectors denote
the basis columns of matrix $\bA$, white columns denote zero vectors, purple entries denote
one, blue and black entries denote elements that are not necessarily zero. The Bayesian ID models get the approximation $\bA\approx\bX\bY$ and the post processing procedure obtains the approximation $\bA\approx\bX\bY\approx\bC\bW$.}
\label{fig:id-column}
\end{figure*}

In this context, the ID of underlying matrices captures our interest.
The ID of an $M\times N$ data matrix $\bA$ can be described by $\bA=\bC\bW+\bP$, where the matrix $\bA$ is approximately factorized into a matrix $\bC\in \real^{M\times K}$ reusing $K$ \textit{basis columns} of $\bA$ (thus $\bC$ is also known as a skeleton of $\bA$) and a matrix $\bW\in \real^{K\times N}$ with entries no greater than 1 in magnitude; 
the error is captured by an $M\times N$ matrix $\bP$. 
Training such models amounts to finding the optimal rank-$K$ approximation to the observed $M\times N$ data matrix $\bA$ under some loss functions. 
Let $\br\in \{0,1\}^N$ be the \textit{state vector} with each entry indicating the type of the corresponding column, i.e., \textit{basis column} or \textit{interpolated (remaining) column}: if $r_n=1$, then the $n$-th column of $\bA$ is a basis column; on the contrary, the $n$-th column 
is interpolated using the basis columns within a tolerance of error.
Suppose further the set $I$ contains the indices of the interpolated columns with $r_n=0$
and the set $J$ contains the indices of the basis columns with $r_n=1$ 
where 
$$
J\cap I =\emptyset  ; \gap J \cup I =\{1,2,\ldots, N\}.
$$
Then $\bC$ can be described by the Matlab-style notation as $\bC=\bA[:,J]$ where the colon operator implies all indices. The approximation $\bA\approx \bC\bW$ can be equivalently stated that $\bA\approx\bC\bW=\bX\bY$ where $\bX\in \real^{M\times N}$ and $\bY\in \real^{N\times N}$ with 
$$
\begin{aligned}
\bX[:,J]&=\bC\in \real^{M\times K}; \gap &\bX[:,I] &= \bzero\in \real^{M\times (N-K)};\\
\bY[J,:]&=\bW \in \real^{K\times N}; \gap &\bY[I,:] &= \text{random matrix} \in \real^{(N-K)\times N}.
\end{aligned}
$$  
We also notice that there exists a $K\times K$ identity matrix $\bI$ inside $\bW$ and $\bY$:
\begin{equation}\label{equation:submatrix_bid_identity}
	\bI = \bW[:,J] = \bY[J,J].
\end{equation}
Having the equivalence of $\bC\bW=\bX\bY$, the problem of $\bA\approx\bC\bW$ can be stated as finding the approximation $\bA\approx\bX\bY$ alternatively with the state vector $\br$ recovering the submatrix $\bC$ (Figure~\ref{fig:id-column}).
Mean squared error (MSE) is applied to evaluate the \textit{reconstruction error}:
\begin{equation}\label{equation:idbid-per-example-loss}
	\mathop{\min}_{\bW,\bZ} \,\, \frac{1}{MN}\sum_{n=1}^N \sum_{m=1}^{M} \left(a_{mn} - \bx_m^\top\by_n\right)^2,
\end{equation}
where $a_{mn}$ is the $(m,n)$-th element of matrix $\bA$, and $\bx_m$, $\by_n$ are the $m$-th \textbf{row} and $n$-th \textbf{column} of $\bX$, $\bY$ respectively for simplicity.
The magnitude constraint in $\bY$ or $\bW$ is approached by considering the Bayesian ID model as a latent factor model where we employ Bayesian inference to find the latent components via the specified graphical model.
Therefore, no explicit magnitude constraints are considered. 


In this paper, we introduce a novel Bayesian ID (BID) approach with each column of the observed matrix having its score measuring the importance in the model; the larger the score, the higher the priority to select; hence the name \textit{intervened interpolative decomposition (IID)}.
The rest of the paper is organized as follows. We will introduce the vanilla Bayesian ID method in Section~\ref{section:related_iid_word}. Section~\ref{section:iid_main} then presents the proposed IID method. Section~\ref{section:iid_quantaprob} provides one of the applications for the IID method in finding quantitative strategies, followed by the experiments in Section~\ref{section:iid_experiments}.


\section{Related Work}\label{section:related_iid_word}

%
%
%
%
%

\subsection{Bayesian GBT Model for Interpolative Decomposition}

\begin{SCfigure}
\centering
\vspace{-0.55cm} 
\subfigtopskip=2pt 
\subfigbottomskip=6pt 
\subfigcapskip=-2pt 
\includegraphics[width=0.36\textwidth]{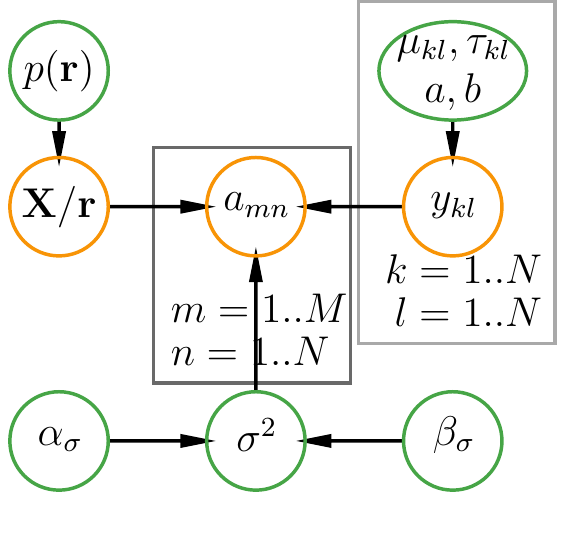}
\caption{Graphical representation of the GBT model where green circles denote prior variables, orange circles represent observed and latent variables, and plates represent repeated variables. Comma ``," in the cycles represents ``and", and ``/" in the cycles represents ``or". Parameters $a,b$ are fixed with $a=-1,b=1$ in our case; while a weaker construction can set them to $a=-2,b=2$.}
\label{fig:bmf_bids_foriid}
\end{SCfigure}

In this section, we review the Bayesian approach for computing the interpolative decomposition.
We consider the data matrix $\bA$ to be generated via the probabilistic generative process (Figure~\ref{fig:bmf_bids_foriid}). The element $a_{mn}$ of matrix $\bA$ is modeled via a Gaussian likelihood function,
\begin{equation}\label{equation:iid_data_entry_likelihood}
	p(a_{mn} | \bx_m^\top\by_n, \sigma^2) = \normal(a_{mn}|\bx_m^\top\by_n, \sigma^2), 
\end{equation}
where $\bx_m^\top\by_n$ and $\sigma^2$ are mean and variance respectively.
Then, we place an inverse-Gamma prior over the data variance (a conjugate prior), 
\begin{equation}\label{equation:prior_iid_gamma_on_variance}
	p(\sigma^2 | \alpha_\sigma, \beta_\sigma) = \inversegammadist(\sigma^2 | \alpha_\sigma, \beta_\sigma),
\end{equation}
where $\inversegammadist(x|\alpha_\sigma, \beta_\sigma)= \frac{(\beta_\sigma)^\alpha}{\Gamma(\alpha_\sigma)} x^{-\alpha_\sigma-1}\exp\{-\frac{\beta_\sigma}{x}\}u(x)$ is an inverse-Gamma density with $\Gamma(\cdot)$ being the gamma function and $u(x)$ being the unit step function that has a value of $1$ when $x\geq0$ and 0 otherwise.

We treat the latent variables $y_{kl}$'s (with $k,l\in \{1,2,\ldots,N\}$, see Figure~\ref{fig:bmf_bids_foriid}) as random variables. 
And in order to express beliefs about the values of these latent variables, we need prior densities over them, for example, a constraint with magnitude smaller than 1, even when there are many additional constraints
(e.g., semi-nonnegativity in \citet{ding2008convex}, nonnegativity in \citet{lu2022flexible, lu2022robust}, or discreteness in \citet{gopalan2014bayesian, gopalan2015scalable}).
Here we assume further that the latent variable $y_{kl}$'s are independently drawn from a general-truncated-normal prior:
\begin{equation}\label{equation:rn_prior_bidd}
	p(y_{kl} | \cdot ) = \generaltruncatednormal(y_{kl} | \mu_{kl}, (\tau_{kl})^{-1}, a=-1, b=1),
\end{equation}
where $\generaltruncatednormal(x|\mu, \frac{1}{\tau}, a, b)=$ $\frac{\sqrt{\frac{\tau}{2\pi}} \exp \{-\frac{\tau}{2}(x-\mu)^2  \}  }{\Phi((b-\mu)\cdot \sqrt{\tau})-\Phi((a-\mu)\cdot \sqrt{\tau})}$$u(x|a,b)$
is a general-truncated-normal (GTN) with zero density below $x=a$ or above $x=b$ and renormalized to integrate to one, $u(x|a,b)$ is a step function that has a value of 1 when $a\leq x\leq b$ and 0 otherwise, and $\Phi(\cdot)$ function is the cumulative distribution function of standard normal density $\normal(0,1)$. The parameters $\mu$ and $\tau$ in GTN are known as the ``parent mean" and ``parent precision" of the original normal distribution $\normal(\mu, \frac{1}{\tau})$. 
This GTN prior is thus utilized to enforce the constraint on the components $\bY$ (or $\bW$) with no entry of $\bY$ having an absolute value greater than 1, and is conjugate to the Gaussian likelihood. 

We call the Bayesian ID method discussed above \textit{GBT} where \textit{G} stands for Gaussian density, \textit{B} stands for Beta-Bernoulli density intrinsically, and \textit{T} is short for general-truncated-normal density.

\paragraph{Hierarchical prior and automatic relevance determination (ARD)}
There is also a hierarchical model on Bayesian inference for ID where we place a joint hyperprior over the hyperparameters $\{\mu_{kl}, \tau_{kl}\}$ of GTN density in Eq.~\eqref{equation:rn_prior_bidd}, i.e., the GTN-scaled-normal-Gamma (GTNSNG) density that can decouple the parameters $\mu_{kl}, \tau_{kl}$, and as a result, their posterior conditional densities are normal and Gamma respectively \citep{lu2022bayesian}.
And also the ARD method can determine the number of columns inside the factored component $\bX$ automatically by a special prior on the state vector $\br$ \citep{lu2022comparative}.
The development of the IID method on the non-hierarchical, hierarchical, and ARD models are the same, and we shall only discuss the non-hierarchical and non-ARD versions for simplicity.

\paragraph{Post processing} The last step shown in Figure~\ref{fig:id-column} presents a step of post processing, where we enforce the identity submatrix in $\bW$ (Eq.~\eqref{equation:submatrix_bid_identity}). This normally can reduce the MSE to a minor extent \citep{lu2022bayesian}. 

\subsection{Gibbs Sampler for GBT Model}
In this section, we only shortly describe the posterior conditional density for Gibbs sampling to find the Bayesian inference. While a step-by-step derivation is provided in \citet{lu2022bayesian, lu2022comparative} for both hierarchical, non-hierarchical, ARD, and non-ARD versions.

Denote all elements of $\bY$ except $y_{kl}$ as $\bY_{-kl}$, 
the conditional density of $y_{kl}$ is also a GTN density and it can be obtained by 
\begin{equation}\label{equation:posterior_gbt_ykl}
\begin{aligned}
&\gap p(y_{kl} | \bA, \bX, \bY_{-kl}, \mu_{kl}, \tau_{kl}, \sigma^2) \propto  p(\bA|\bX,\bY, \sigma^2) \cdot p(y_{kl}|\mu_{kl}, \tau_{kl} )\\
&
\propto \generaltruncatednormal(y_{kl}| \widetilde{\mu},( \widetilde{\tau})^{-1}, a=-1,b=1),
\end{aligned}
\end{equation}
where $\widetilde{\tau} =\frac{\sum_{i}^{M}  x_{ik} ^2}{\sigma^2} +\tau_{kl}$ is the posterior ``parent precision" of the GTN distribution, and 
$
\widetilde{\mu} = \big(\frac{1}{\sigma^2}  \sum_{i}^{M} x_{ik}  \big(a_{il}-\sum_{j\neq k}^{N}x_{ij}
y_{jl}\big)
+\textcolor{black}{\tau_{kl}\mu_{kl}}
\big) \big/ \widetilde{\tau}
$
is the posterior ``parent mean" of the GTN distribution.

Given the state vector $\br=[r_1,r_2, \ldots, r_N]^\top\in \real^N$ such that the index set $J = J(\br) = \{n|r_n = 1\}_{n=1}^N$ and $I = I(\br) = \{n|r_n = 0\}_{n=1}^N$. To draw a state vector $\br$, we can select one index $j\in J$ and another index $i\in I$ (where the old values are $r_j=1$ and $r_i=0$) such that 
\begin{equation}\label{equation:postrerior_gbt_rvector}
\begin{aligned}
o_j &= 
\frac{p(r_j=0, r_i=1|\bA,\sigma^2, \bY, \br_{-ji})}
{p(r_j=1, r_i=0|\bA,\sigma^2, \bY, \br_{-ji})}\\
&=
\frac{p(r_j=0, r_i=1)}{p(r_j=1, r_i=0)} \times
\frac{p(\bA|\sigma^2, \bY, \br_{-ji}, r_j=0, r_i=1)}{p(\bA|\sigma^2, \bY, \br_{-ji}, r_j=1, r_i=0)},
\end{aligned}
\end{equation}
where $\br_{-ji}$ denotes all elements of $\br$ except the $j$-th and $i$-th entries.
In GBT, every column has same priority so we have $p(r_j=0, r_i=1)=p(r_j=1, r_i=0)$. Then the conditional probability of $p(r_j=0, r_i=1|\bA,\sigma^2, \bY, \br_{-ji})$ can be obtained by 
\begin{equation}\label{equation:postrerior_gbt_rvector222}
p(r_j=0, r_i=1|\bA,\sigma^2, \bY, \br_{-ji}) = \frac{o_j}{1+o_j}.
\end{equation}

Finally, by conjugacy, the conditional posterior density of $\sigma^2$ is an inverse-Gamma distribution:
\begin{equation}\label{equation:posterior_gnt_sigma2}
	\begin{aligned}
		&\gap p(\sigma^2 | \bX, \bY, \bA)
		= \inversegammadist(\sigma^2 | \widetilde{\alpha_\sigma}, \widetilde{\beta_\sigma}),
	\end{aligned}
\end{equation}
where $\widetilde{\alpha_\sigma} = \frac{MN}{2}+\alpha_\sigma$, 
$\widetilde{\beta_\sigma}=\frac{1}{2} \sum_{i,j=1}^{M,N}(a_{ij}-\bx_i^\top\by_j)^2+\beta_\sigma$ are the posterior shape and scale parameters for the inverse-Gamma density. The procedure for GBT is then formulated in Algorithm~\ref{alg:gbt_iid_gibbs_sampler}.

\begin{algorithm}[!htb] 
\caption{Gibbs sampler for GBT (and IID) model. The procedure presented here can be inefficient but is explanatory. While a vectorized manner can be implemented to find a more efficient algorithm. By default, weak priors are $\alpha_\sigma=0.1, \beta_\sigma=1$, $\{\mu_{kl}\}=0, \{\tau_{kl}\}=1$, and $a=-1, b=1$ are fixed. Set the latent dimension $K$.} 
\label{alg:gbt_iid_gibbs_sampler}  
\begin{algorithmic}[1] 
\FOR{$t=1$ to $T$}
\STATE Sample state vector $\br$ from Eq.~\eqref{equation:postrerior_gbt_rvector222} (based on Eq.~\eqref{equation:postrerior_gbt_rvector} for GBT, or on Eq.~\eqref{equation:posterior_IID} for IID);
\STATE Update matrix $\bX$ by $\bA[:,J]$ where index vector $J$ is the index of $\br$ with value 1 and set $\bX[:,I]=\bzero$ where index vector $I$ is the index of $\br$ with value 0;
\STATE Sample variance $\sigma^2$ from $p(\sigma^2 | \bX,\bY, \bA)$ in Eq.~\eqref{equation:posterior_gnt_sigma2}; 
\FOR{$k=1$ to $N$} 
\FOR{$l=1$ to $N$} 
\STATE Sample factored component $y_{kl}$ from $p(y_{kl} | \bA, \bX, \bY_{-kl}, \mu_{kl}, \tau_{kl}, \sigma^2)$ in Eq.~\eqref{equation:posterior_gbt_ykl};
\ENDFOR
\ENDFOR
\STATE Output $||\bA-\bX\bY||_2$ loss in Eq.~\eqref{equation:idbid-per-example-loss}, and stop iteration if it converges.
\ENDFOR
\STATE Output mean loss in Eq.~\eqref{equation:idbid-per-example-loss} for evaluation after burn-in iterations.
\end{algorithmic} 
\end{algorithm}

\section{Intervened Interpolative Decomposition (IID)}\label{section:iid_main}
Going further from the GBT model, we propose the intervened interpolative decomposition (IID) algorithm.
The proposed IID algorithm has exactly the same 
generative process as shown in Eq.~\eqref{equation:iid_data_entry_likelihood},
inverse-Gamma prior on the variance parameter $\sigma^2$ in Eq.~\eqref{equation:prior_iid_gamma_on_variance},
and GTN prior over the latent variables $y_{kl}$'s in Eq.~\eqref{equation:rn_prior_bidd}.
However, we consider further that some columns of the observed matrix $\bA$ has a larger importance  that \textit{should} be selected with a higher priority over the other columns.

Suppose the importance of each column of the observed matrix $\bA$ is captured by a \textit{raw importance vector} $\widehat{\bp}\in \real^N$
where $\widehat{p}_i \in [-\infty, \infty]$ for all $i$ in $\{1,2,\ldots, N\}$. The raw importance vector can then be transformed into the range 0 to 1
$$
\bp = \text{Sigmoid}(\widehat{\bp}),
$$
where \textit{Sigmoid($\cdot$)} is the $f(x) = \frac{1}{1+\exp\{-x\}}$ that can return value in the range 0 to 1.
The Sigmoid function acts as a squashing function because its domain is the set of all real numbers, and its range is (0, 1).
Then  we take the $\bp$ vector as the final \textit{importance vector} to indicate the importance of each column in the matrix $\bA$.

Going further from Eq.~\eqref{equation:postrerior_gbt_rvector}, the intermediate variable $o_j$ is calculated instead by 
\begin{equation}\label{equation:posterior_IID}
\begin{aligned}
o_j 
&=
\frac{p(r_j=0, r_i=1)}{p(r_j=1, r_i=0)} 
\times
\frac{p(\bA|\sigma^2, \bY, \br_{-ji}, r_j=0, r_i=1)}{p(\bA|\sigma^2, \bY, \br_{-ji}, r_j=1, r_i=0)}\\
&=
\textcolor{blue}{
	\frac{1-p_j }{p_j}
	\frac{p_i }{1-p_i}
}
\times 
\frac{p(\bA|\sigma^2, \bY, \br_{-ji}, r_j=0, r_i=1)}{p(\bA|\sigma^2, \bY, \br_{-ji}, r_j=1, r_i=0)}.
\end{aligned}
\end{equation}
And again, the conditional probability of $p(r_j=0, r_i=1|\bA,\sigma^2, \bY, \br_{-ji})$ can be obtained by 
\begin{equation}\label{equation:postrerior_gbt_rvector222_IID}
	p(r_j=0, r_i=1|\bA,\sigma^2, \bY, \br_{-ji}) = \frac{o_j}{1+o_j}.
\end{equation}
Since we intervene in the procedure of the Gibbs sampling in Eq.~\eqref{equation:posterior_IID}, hence the name \textit{intervened interpolative decomposition (IID)}.

%


\section{Quantitative Problem Statement}\label{section:iid_quantaprob}
After developing the intervened interpolative decomposition algorithm, one may get confused about why it is so important and curious about the applications it can be applied in practice.
It is well known that large quantitative hedge funds and asset managers have been recruiting a large number of data miners and financial engineers in order to build effective alphas, and the number of alpha components might climb into the millions or perhaps billions \citep{tulchinsky2019finding}. 
As a result, creating a meta-alpha from all of the alphas or a large fraction of the alpha pool might be troublesome for the following reasons: 
a). If we use the same alphas as others, some illiquid alphas with low volume will be traded heavily. This will make the strategy meaningless due to capacity constraints; 
b). Using too many alphas may result in overfitting, resulting in poor out-of-sample (OS) performance;
c). Many alphas might be mutually dependent, and certain machine learning algorithms, such as neural networks, might uncover their limits caused by multi-linear difficulties while attempting to determine the meta-strategy from the entire set of alphas;
d). Finding trading signals from the full alpha pool can be time-consuming because of limited computing resources;
e). To minimize market risks, we constantly aim to discover a distinct subset of alphas to test alternative methods with low correlation.
For the five reasons stated above, there is an urgent need to design algorithms that choose a small subset of alphas from a large pool of them in order to prevent overfitting, make the final approach more scalable, and obtain the findings in a reasonable amount of time.
It is trivial to select an appropriate subset by the \textit{RankIC} metric (see definition below), i.e., we select the alphas having the highest RankIC values. However, the problems still remain that the selected subset will not represent the whole pool of alphas, and the selected alphas may be mutually dependent.

Our objective is to identify as many representative alpha factors as possible with optimal performance. The selected subset of alphas is representative in the sense that the small subset of alphas can be used to reconstruct other alphas with a small replication error.
The traditional ID algorithm, either using a \textit{Randomized algorithm} \citep{liberty2007randomized} or a Bayesian approach we have discussed above, can only help to find the representative ones. However, the end choices may seem to select alphas with low performance.
Using the proposed IID method, on the other hand, can help find the representative (that can reconstruct other alphas with small error) and the desirable (high RankIC scores) alphas at the same time.

\subsection{Formulaic Alphas}
WorldQuant, a quantitative investment management firm, previously disclosed 101 formulaic short-term alpha determinants in 2016 \citep{kakushadze2016101}. Since then, the 191 alpha factors from Guotai Junan Securities \citep{guotaijunan2017} have also been welcomed by many investors and institutions.
These formulaic alpha components are derived from several stock data elements, including, among others, volumes, prices, volatilities, and volume-weighted average prices (vwap).
As the name implies, a formulaic alpha is a type of alpha that can be expressed as a formula or a mathematical expression. For example, a \textit{mean-reversion} alpha can be expressed in terms of a mathematical expression as follows:
$$
\text{Alpha =  }- \left( \text{close(today) $-$close(5$\_$days$\_$ago ) } \right)/ 
\text{close(5$\_$days$\_$ago)}.
$$
In this sense, we take the opposite action as indicated by the closing price: we go short if the price has risen during the previous five days, and we go long otherwise. At a high level, the alpha value indicates the trend of the price in the days to come; the higher the alpha value for each stock, the more likely it is that the stock's price will rise in the next few days.

\subsection{Evaluation Metrics}
Let $r_{t}$ denote the yield rate of stock $s$ on $t$-th day. Suppose further 
$p_t$ is asset closing price at time $t$ where $t\in \{1,2,\ldots, T\}$, the
return of the asset at time $t$ can be obtained by the following
equation:
\begin{equation}
r_t = \frac{p_t - p_{t-1}}{p_t}.
\end{equation}
We use the Rank information coefficient (\textit{RankIC}) to evaluate the effectiveness of an alpha:
\begin{equation}
\text{RankIC}(\ba, \br^h)=\text{Spearman}(\ba, \br^h),
\end{equation}
where $\text{Spearman}(\cdot)$ indicates the Spearman correlation, $\ba$ is the sequence of an alpha, $\br^h$ is the sequence of the return value with holding period $h$ such that the $i$-th element of $\br^h$ represent the daily return of $h$ days later. The RankIC then can be used as an indicator of the importance of each alpha factor and plugged into Eq.~\eqref{equation:posterior_IID} directly.


\section{Experiments}\label{section:iid_experiments}

For each stock $s$  (i.e., $s\in \{1,2,\ldots, S\}$ where $S$ is the total number of stocks), we have a matrix $\bA_s$ with shape $\bA_s\in \real^{N\times D}$ where $N$ is the number of alphas and $D$ is the number of dates so that each row of $\bA_s$ is regarded as an alpha series. We want to select a subset of the alphas (here we assume $M$ out of the $N$ alphas are selected). The RankIC between each alpha series and the delayed return series with horizon $h=1$ is then taken as the \textit{important value} directly, a higher RankIC indicates a  higher priority.

\begin{table}[h]
\centering
\small 
\setlength{\tabcolsep}{7.4pt}
\begin{tabular}{llllll}
\hline
Ticker & Type & Sector & Company &  Average Amount \\ \hline
SH601988   & Share & Bank               & Bank of China Limited    &  427,647,786  \\
SH601601   & Share & Public Utility    & China Pacific Insurance (Group)   & 819,382,926 \\
SH600028    & Share & Public Utility & China Petroleum \& Chemical Corporation              &  748,927,952\\
SH600016    & Share & Bank    & China Minsheng Banking Corporation &285,852,414  \\
SH601186    & Share & Public Utility    & China Railway Construction Corporation  & 594,970,588\\
SH601328    & Share & Bank & Bank of Communications Corporation &  484,445,915  \\
SH601628     & Share & Public Utility & China Life Insurance Company Limited&368,179,861  \\
SH601939    & Share   & Bank      & China Construction Bank Corporation &527,876,669  \\
\hline
SH510300    & ETF   & CSI 300         & Huatai-PineBridge CSI 300 ETF  &1,960,687,059  \\
SH510050      & ETF   & CSI 50 & ChinaAMC China CSI 50 ETF    &2,020,385,879  \\
\hline
\end{tabular}
	\vspace{-0.25cm} 
\caption{Summary of the underlying portfolios in the China market, ten assets in total. The average amount (in the currency of RMB) is calculated in the period of the test set.}
\label{table:iid_cn_data_summary}		
\end{table}

\begin{figure*}[h]
\centering  
\vspace{-0.3cm} 
\subfigtopskip=2pt 
\subfigbottomskip=9pt 
\subfigcapskip=-5pt 
\subfigure[Ten different portfolios where we initialize each portfolio with a unitary value for clarity.]{\includegraphics[width=0.49\textwidth]{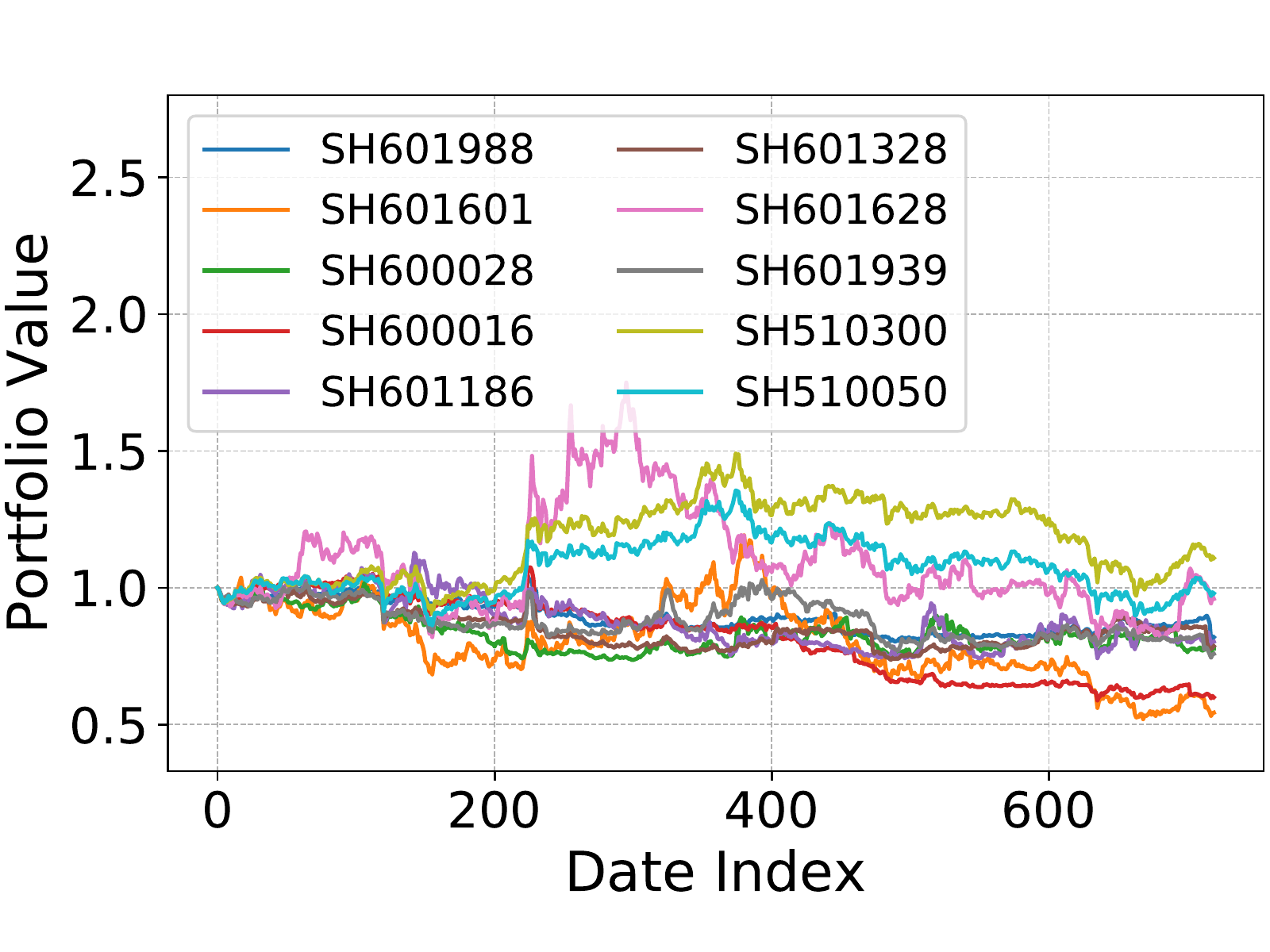} \label{fig:bid_iid_datasets_ashare}}
\subfigure[Portfolio values with the same strategy by using different alphas via comparative selection models.]{\includegraphics[width=0.49\textwidth]{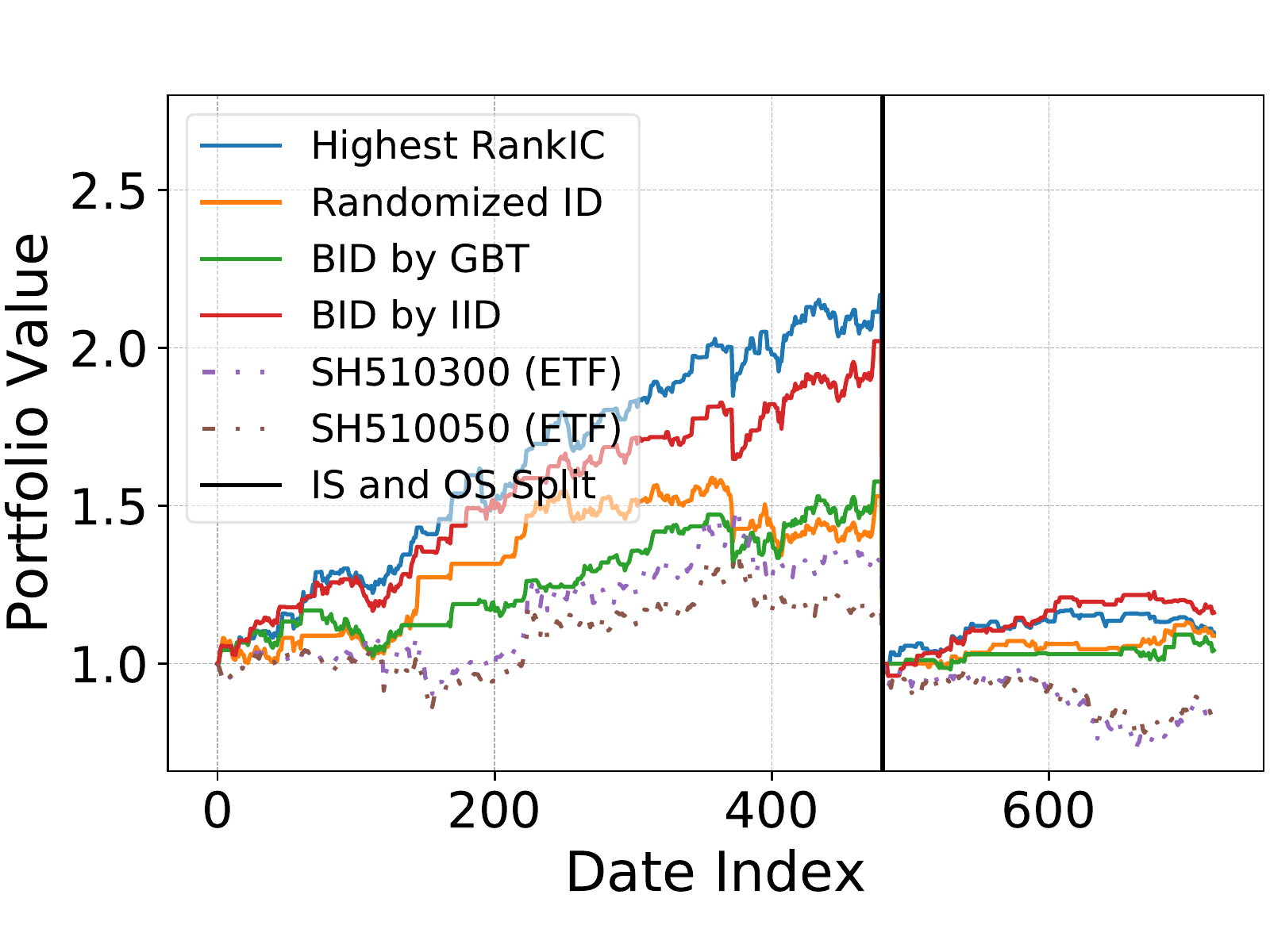} \label{fig:bid_iid_portfolio_ashare}}
\vspace{-0.7cm} 
\caption{Portfolio values of the ten assets (left), and the portfolio values (right) of different methods where we split by in-sample and out-of-sample periods, and initialize with a unitary value for each period. The proposed IID performs better in the out-of-sample period (see also Table~\ref{table:iid_selected_mean_ic}).}
\label{fig:bid_iid_portfolio_ashare_full}
\end{figure*}

\paragraph{Dataset}
To assess the proposed algorithm and highlight the primary benefits of the IID technique, we perform experiments with several analytical tasks and use data for ten assets from the China market and diverse industrial areas, including Bank, Public Utility, and ETF.
We obtain publicly available data from tushare \footnote{\url{https://tushare.pro/}.}. 
The data covers a three-year period, i.e., 2018-07-18 to 2021-07-05 (720 trading days), where the data between 2018-07-18 and 2020-07-09 is considered the training set (480 calendar days); 
while data between 2020-07-10 and 2021-07-05 is taken as the test set (240 trading days).
The underlying portfolios are summarized in Table~\ref{table:iid_cn_data_summary} and Figure~\ref{fig:bid_iid_datasets_ashare} shows the series of different assets where
we initialize each portfolio with a unitary value for clarity.
The assets are chosen by selecting the ones with high amount values (random ten assets among the fifty assets with highest average amounts in China market during the selected period) so that there are fewer trading restrictions.

We obtain 78 alphas from the 101 formulaic alphas \citep{kakushadze2016101}, 94 alphas from the 191 formulaic alphas \citep{guotaijunan2017}, and 19 proprietary alphas. The alphas are chosen to have a value that is neither too large nor too small.
In this sense, the alpha matrix $\bA_s$ is of shape $214\times 480$ for each asset.  

In all scenarios, the same parameter initialization is adopted when conducting different tasks. 
Experimental evidence demonstrates that post-processing can marginally improve performance.
For clarification, we only provide the findings of the GBT and IID models after post processing.
The IID model can select the important features (alphas) with a higher priority while keeping the reconstructive error as small as possible, resulting in performance that is as good as or better than the vanilla GBT method in low-rank ID approximation across a wide range of experiments on different datasets.

We use mean squared error (MSE, Eq.~\eqref{equation:idbid-per-example-loss}), which measures the similarity between the observed and reconstructive matrices, to evaluate the overall decomposition performance; the smaller the value, the better the performance.

\begin{figure*}[h]
\centering  
\vspace{-0.2cm} 
\subfigtopskip=2pt 
\subfigbottomskip=0pt 
\subfigcapskip=-2pt 
\subfigure[Convergence of the models on the 
SH510050, SH510300, SH601939, SH601628, and SH601328 datasets, as measured by MSE. The algorithm almost converges in less than 100 iterations.]{\includegraphics[width=1\textwidth]{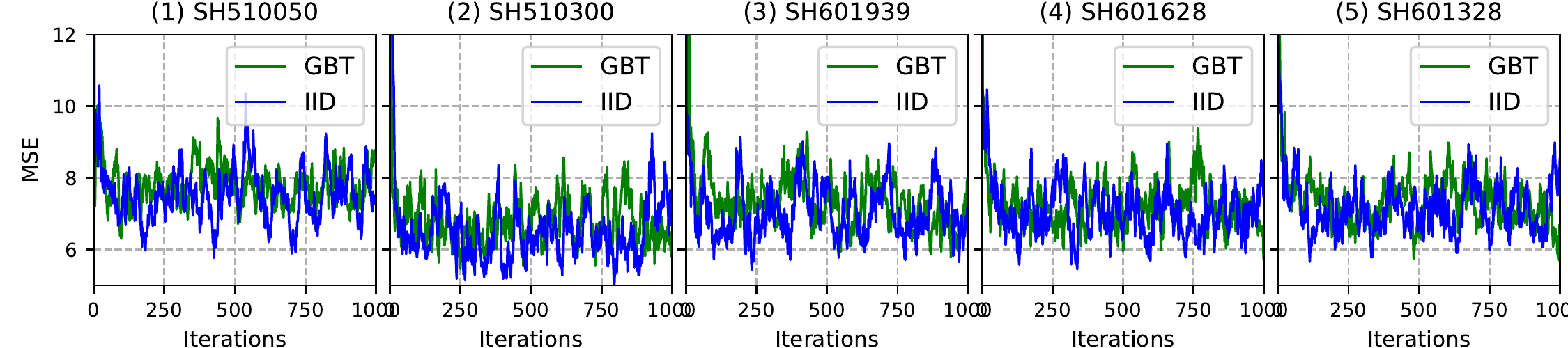} \label{fig:iid_alpha_convergence}}
\subfigure[Averaged autocorrelation coefficients of samples of $y_{kl}$ computed using Gibbs sampling on the 
SH510050, SH510300, SH601939, SH601628, and SH601328 datasets.]{\includegraphics[width=1\textwidth]{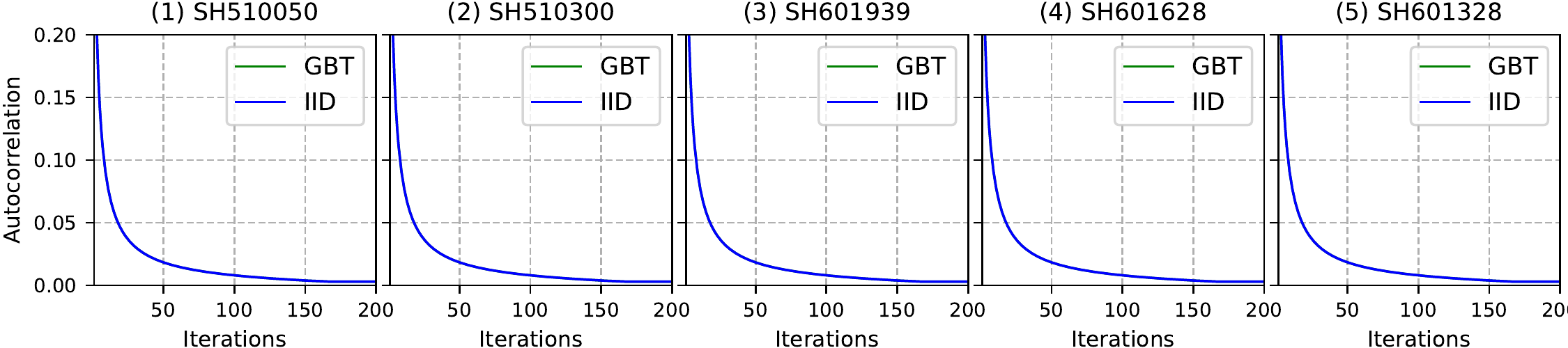} \label{fig:iid_alpha_autocorrelation}}
\vspace{-0.3cm} 
\caption{Convergence results (upper), and sampling mixing analysis (lower) on the SH510050, SH510300, SH601939, SH601628, and SH601328 datasets for a latent dimension of $K=10$.
}
\label{fig:allresults_bids_ard}
\end{figure*}
\paragraph{Hyperparameters}
In those experiments, we use  $a=-1, b=1,\alpha_\sigma=0.1, \beta_\sigma=1$, ($\{\mu_{kl}\}=0, \{\tau_{kl}\}=1$) for both GBT and IID models.
The adopted parameters are  uninformative and weak prior choices and the models are insensitive to them.
The observed or unobserved variables are initialized from random draws as long as those hyperparameters are fixed since this initialization method provides a better initial guess of the correct patterns in the matrices.
In all cases, we execute 1,000 iterations of Gibbs sampling with a burn-in of 100 iterations and a thinning of 5 iterations, since the convergence analysis indicates the algorithm can converge in fewer than 100 iterations.

\begin{table}[]
\centering
\vspace{-0.35cm} 
\scriptsize
\setlength{\tabcolsep}{3pt}
\begin{tabular}{lllllllllll}
\hline 
 & SH601988 & SH601601 & SH600028 & SH600016 & SH601186 & SH601328 & SH601628 & SH601939 & SH510300 & SH510050\\
 \hline 
GBT Min & 5.235 & 5.814 & 5.235 & \textbf{6.381}& 5.819 & 5.700 & 5.734 & 5.785 & 5.462 & 6.297 \\
IID Min & \textbf{4.567} &\textbf{5.700}& \textbf{4.843} & 6.490 & \textbf{5.104} & \textbf{5.658} & \textbf{5.445} & \textbf{5.435} & \textbf{4.876} & \textbf{5.767} \\
GBT Mean & 6.476 & \textbf{7.367} & 6.764 & 8.053 & 7.066 & 7.250 & 7.206 & 7.242 & 6.769 & 7.776 \\
IID Mean & \textbf{6.239} & 7.449 & \textbf{6.664} & \textbf{7.831} & \textbf{6.558} & \textbf{7.081} & \textbf{7.002} & \textbf{7.031} & \textbf{6.450} & \textbf{7.492} \\
\hline
\end{tabular}
\vspace{-0.3cm} 
\caption{Minimal and mean MSE measures after burn-in across different iterations for GBT and IID models on the 10 alpha matrices from 10 assets. In all cases, $K=10$ is set as the latent dimension. 
In most cases, the results of IID converge to a smaller value than the GBT model.}
\label{table:comparis_gbt_iid_mse}
\vspace{-0.1cm} 
\end{table}

\subsection{Convergence and Comparative Analysis}
We first show the rate of convergence over iterations on different assets. Due to space constraints, we omit convergence results for the first five assets and only present those for portfolios SH510050, SH510300, SH601939, SH601628, and SH1303. Results for the other assets are qualitatively similar.

We run GBT and IID models with $K=10$ for the five datasets where $214$ is the full rank of the matrices, and the error is measured by MSE.
Figure~\ref{fig:iid_alpha_convergence} shows the rate of convergence over iterations. Figure~\ref{fig:iid_alpha_autocorrelation} shows autocorrelation coefficients of samples computed using Gibbs sampling. We observe that the mixings of the IID are close to those of GBT. When the lags are greater than ten, the coefficients are less than 0.1, indicating that the Gibbs sampler mixes well. 
In all experiments, the algorithm converges in less than 100 iterations.
We also observe that the IID model does not converge to a larger error than the vanilla GBT model, though we put more emphasis on selecting the columns with high RankIC. Table~\ref{table:comparis_gbt_iid_mse} presents the minimal MSE and mean MSE after burn-in across different iterations for GBT and IID models on the ten alpha matrices from ten assets. In most cases, the IID can even converge to a smaller MSE value.

\begin{algorithm}[!htb] 
\caption{Alpha selection for portfolio allocation. Gibbs sampler for GBT and GBTN ID models. 
Select holding period $h$, number of alphas to select $M$.
} 
\label{alg:iid_alpha_selection}  
\begin{algorithmic}[1] 
\STATE Split the alpha matrix for in-sample (IS) and out-of-sample (OS) evaluations: 
$$
\bA_{\text{in}} = \bA_s[:,0:D_{\text{in}}] \in \real^{N\times D_{\text{in}}}, \gap 
\bA_{\text{out}} = \bA_s[:,D_{\text{in}}+1:D]\in \real^{N\times (D-D_{\text{in}})};
$$
\STATE Using ID to decide the alphas to be selected  on matrix $\bA_{\text{in}}^\top$, with the selected indices $\bmm$:
$$
\widehat{\bA}_{\text{in}} = \bA_s[\bmm,0:D_{\text{in}}] \in \real^{\textcolor{blue}{M}\times D_{\text{in}}}, \gap 
\widehat{\bA}_{\text{out}} = \bA_s[\bmm,D_{\text{in}}+1:D]\in \real^{\textcolor{blue}{M}\times (D-D_{\text{in}})};
$$
\FOR{$m=1$ to $M$}
\STATE Using the $m$-th IS alpha vector $\ba_m=\widehat{\bA}_{\text{in}}[m,:]\in \real^{D_{\text{in}}}$ to decide the weight $\bw$ and interception $b$ via ordinary least squares (OLS) so that the MSE between the prediction $\ba_m^\top\bw_m +b_m$ and the shifted return vector $\br^h$ is minimized, i.e., minimizing $\text{MSE}(\ba_m^\top\bw_m +b_m, \br^h)$. The weight and interception are then used in OS evaluation.
\ENDFOR
\FOR{$d=1$ to $D-D_{\text{in}}$}
\STATE On each day in the OS period, we use the mean evaluation of each prediction from the $M$ alphas to decide to go long or not, i.e., to go long if $\sum_{m=1}^M \ba_m^\top\bw_m +b_m >0$; and do nothing otherwise since we restrict the analysis to  long-only portfolios.
%
Though we employ a long-only portfolio, we can favor a \textit{market-neutral strategy}: we open long positions only when we anticipate that at least half of the stocks will rise on the following $h$ day, and we weight each stock equally.

\ENDFOR
\end{algorithmic} 
\end{algorithm}

\begin{table}[]
\centering
\vspace{-0.4cm} 
\setlength{\tabcolsep}{10pt}
\small
\begin{tabular}{lllll}
\hline 
Methods & Highest RankIC & Randomized ID & BID with GBT & BID with IID \\
\hline
Mean RankIC & \textbf{0.1035}         & 0.0651        & 0.0553       & \textbf{0.0752}  \\
Mean Correlation & 0.2276$\downarrow$         & 0.5741$\downarrow$        & \textbf{0.1132}       & \textbf{0.1497}  \\
\hline
Sharpe Ratio (OS) & 1.0276 & 1.0544 & 0.5045 & \textbf{1.5721}  \\
Sharpe Ratio (IS) & \textbf{2.6511} & 1.3019 & 1.4965 & 2.3231  \\
\hline
Annual Return (OS) & 0.1043 & 0.0932 & 0.0484 & \textbf{0.1633}  \\
Annual Return (IS) & \textbf{0.4390} & 0.2281 & 0.2425 & 0.3805  \\
\hline
Max Drawdown (OS) & 0.0632 & \textbf{0.0373} & \textbf{0.0484} & \textbf{0.0552}  \\
Max Drawdown (IS) & \textbf{0.0892} & 0.1548 & 0.1232 & \textbf{0.0975}  \\
\hline
\end{tabular}
\vspace{-0.3cm} 
\caption{Mean RankIC and correlation of the selected alphas across various assets for different methods. A higher mean RankIC and a lower mean correlation are better. The proposed IID method can find the trade-off between the mean RankIC and the mean correlation.
In all cases, \textit{IS} means in-sample measurements, and \textit{OS} means out-of-sample measurements. The symbol ``$\downarrow$" means the performance is extremely poor.  
}
\label{table:iid_selected_mean_ic}
\end{table}

\subsection{Quantitative Strategy}

After executing the GBT and IID algorithms for computing the interpolative decomposition of each asset's alpha matrix, the state vector $\br$ for each asset is saved and the ten alphas with the largest mean selection during the 1,000 iterations are chosen (with a burn-in of 100 iterations, and thinning of 5 iterations).

Then we follow the quantitative strategy in Algorithm~\ref{alg:iid_alpha_selection} (in which case $h=1$, $N=214$ alphas, $M=10$ alphas, $D=720$ trading days, and $D_{\text{in}}=480$ trading days).
The procedure shown in Algorithm~\ref{alg:iid_alpha_selection} is a very simple quantitative strategy. However, the algorithm can show precisely how the proposed IID method can work in practice.

The strategy using the alphas selected by the proposed IID method is only slightly worse than the one selecting the \textit{highest RankIC} alphas for the in-sample (IS) performance in terms of Sharpe ratio, annual return, and maximum drawdown; however, the IID performs better in the out-of-sample (OS) scenario and this is what we actually want (see Table~\ref{table:iid_selected_mean_ic} and Figure~\ref{fig:bid_iid_portfolio_ashare}).
To evaluate the strategy, we also adopt the Randomized algorithm to compute the ID for comparison \citep{liberty2007randomized}, termed \textit{Randomized ID}. The Randomized ID performs even worse than BID with GBT (see Table~\ref{table:iid_selected_mean_ic}).
Though the IID does not select alphas with the highest RankIC values, this does not mean that the alpha selection procedure is meaningless for the following reasons:
1). \textit{Pool size}: We only use a small alpha pool that only contains 214 alpha factors. When the number of alphas is approaching millions or even billions, the alpha selection procedure is expected to work better. 
2). \textit{Correlation}: The mean correlation of selected alphas across the ten assets of the proposed IID method is smaller than the highest RankIC method. In this sense, the alphas of the latter method have high correlations and a low diversity. If the correlated alphas have low liquidity or perform poorly during a given period, the strategy's risk might increase.
3). \textit{Machine learning models}: In our test, we only use OLS to find the weight of each alpha. For more complex models, e.g., neural networks, the correlated alphas can cause multi-linear problems so that the performance and interpretability are hampered.
4). \textit{Diversification}: Even if selecting the alphas with the highest RankIC can work well in practice, we also want to diversify the strategies so that we are not exposed to specific risks.
The proposed IID method can help find different strategies.

\section{Conclusion}
The purpose of this paper is to propose a novel Bayesian identification algorithm that can select the most significant features while still representing the entire feature pool. The proposed IID method is computationally efficient and requires minimal additional processing.
Overall, we demonstrate that the convergence results of the presented IID model are comparable to those of the existing GBT model.
Similar to vanilla GBT, the IID model can ensure numerical stability by restricting the magnitude of the factored matrix to no more than one.


\bibliography{bib}
\bibliographystyle{iclr}

\end{document}

%% file: symbols.tex
\newcommand{\real}{\mathbb{R}}

\newcommand{\gap}{\,\,\,\,\,\,\,\,}

\newcommand{\ba}{\bm{a}}
\newcommand{\bA}{\bm{A}}

\newcommand{\bC}{\bm{C}}

\newcommand{\bI}{\bm{I}}

\newcommand{\bmm}{\bm{m}}

\newcommand{\bp}{\bm{p}}
\newcommand{\bP}{\bm{P}}

\newcommand{\br}{\bm{r}}

\newcommand{\bw}{\bm{w}}
\newcommand{\bW}{\bm{W}}
\newcommand{\bx}{\bm{x}}
\newcommand{\bX}{\bm{X}}
\newcommand{\by}{\bm{y}}
\newcommand{\bY}{\bm{Y}}

\newcommand{\bZ}{\bm{Z}}

\newcommand{\bzero}{\mathbf{0}}

\definecolor{blue-violet}{rgb}{0.54, 0.17, 0.89}
\definecolor{brightlavender}{rgb}{0.44, 0.16, 0.39}

\newcommand{\inversegammadist}{\mathcal{G}^{-1}}

\newcommand{\normal}{\mathcal{N}}

\newcommand{\generaltruncatednormal}{\mathcal{GTN}}

\mathchardef\mhyphen="2D











%% file: BMF-BID-AlphaDiversification.bbl
\begin{thebibliography}{23}
\providecommand{\natexlab}[1]{#1}
\providecommand{\url}[1]{\texttt{#1}}
\expandafter\ifx\csname urlstyle\endcsname\relax
  \providecommand{\doi}[1]{doi: #1}\else
  \providecommand{\doi}{doi: \begingroup \urlstyle{rm}\Url}\fi

\bibitem[Cheng et~al.(2005)Cheng, Gimbutas, Martinsson, and
  Rokhlin]{cheng2005compression}
Hongwei Cheng, Zydrunas Gimbutas, Per-Gunnar Martinsson, and Vladimir Rokhlin.
\newblock On the compression of low rank matrices.
\newblock \emph{SIAM Journal on Scientific Computing}, 26\penalty0
  (4):\penalty0 1389--1404, 2005.

\bibitem[Dash et~al.(2002)Dash, Choi, Scheuermann, and Liu]{dash2002feature}
Manoranjan Dash, Kiseok Choi, Peter Scheuermann, and Huan Liu.
\newblock Feature selection for clustering-a filter solution.
\newblock In \emph{2002 IEEE International Conference on Data Mining, 2002.
  Proceedings.}, pp.\  115--122. IEEE, 2002.

\bibitem[Ding et~al.(2008)Ding, Li, and Jordan]{ding2008convex}
Chris~HQ Ding, Tao Li, and Michael~I Jordan.
\newblock Convex and semi-nonnegative matrix factorizations.
\newblock \emph{IEEE transactions on pattern analysis and machine
  intelligence}, 32\penalty0 (1):\penalty0 45--55, 2008.

\bibitem[Dy \& Brodley(2004)Dy and Brodley]{dy2004feature}
Jennifer~G Dy and Carla~E Brodley.
\newblock Feature selection for unsupervised learning.
\newblock \emph{Journal of machine learning research}, 5\penalty0
  (Aug):\penalty0 845--889, 2004.

\bibitem[Golub et~al.(1987)Golub, Hoffman, and
  Stewart]{golub1987generalization}
Gene~H Golub, Alan Hoffman, and Gilbert~W Stewart.
\newblock A generalization of the {Eckart-Young-Mirsky} matrix approximation
  theorem.
\newblock \emph{Linear Algebra and its applications}, 88:\penalty0 317--327,
  1987.

\bibitem[Gopalan et~al.(2014)Gopalan, Ruiz, Ranganath, and
  Blei]{gopalan2014bayesian}
Prem Gopalan, Francisco~J Ruiz, Rajesh Ranganath, and David Blei.
\newblock Bayesian nonparametric poisson factorization for recommendation
  systems.
\newblock In \emph{Artificial Intelligence and Statistics}, pp.\  275--283.
  PMLR, 2014.

\bibitem[Gopalan et~al.(2015)Gopalan, Hofman, and Blei]{gopalan2015scalable}
Prem Gopalan, Jake~M Hofman, and David~M Blei.
\newblock Scalable recommendation with hierarchical poisson factorization.
\newblock In \emph{UAI}, pp.\  326--335, 2015.

\bibitem[GuotaiJunan(2017)]{guotaijunan2017}
Securities GuotaiJunan.
\newblock Multi factor stock selection system based on the characteristics of
  short cycle price.
\newblock 2017.

\bibitem[Hou et~al.(2011)Hou, Nie, Yi, and Wu]{hou2011feature}
Chenping Hou, Feiping Nie, Dongyun Yi, and Yi~Wu.
\newblock Feature selection via joint embedding learning and sparse regression.
\newblock In \emph{Twenty-Second international joint conference on Artificial
  Intelligence}, 2011.

\bibitem[Kakushadze(2016)]{kakushadze2016101}
Zura Kakushadze.
\newblock 101 formulaic alphas.
\newblock \emph{Wilmott}, 2016\penalty0 (84):\penalty0 72--81, 2016.

\bibitem[Li et~al.(2009)Li, Zhang, and Sindhwani]{li2009non}
Tao Li, Yi~Zhang, and Vikas Sindhwani.
\newblock A non-negative matrix tri-factorization approach to sentiment
  classification with lexical prior knowledge.
\newblock In \emph{Proceedings of the Joint Conference of the 47th Annual
  Meeting of the ACL and the 4th International Joint Conference on Natural
  Language Processing of the AFNLP}, pp.\  244--252, 2009.

\bibitem[Liberty et~al.(2007)Liberty, Woolfe, Martinsson, Rokhlin, and
  Tygert]{liberty2007randomized}
Edo Liberty, Franco Woolfe, Per-Gunnar Martinsson, Vladimir Rokhlin, and Mark
  Tygert.
\newblock Randomized algorithms for the low-rank approximation of matrices.
\newblock \emph{Proceedings of the National Academy of Sciences}, 104\penalty0
  (51):\penalty0 20167--20172, 2007.

\bibitem[Lim \& Teh(2007)Lim and Teh]{lim2007variational}
Yew~Jin Lim and Yee~Whye Teh.
\newblock Variational {B}ayesian approach to movie rating prediction.
\newblock In \emph{Proceedings of KDD cup and workshop}, volume~7, pp.\
  15--21. Citeseer, 2007.

\bibitem[Lu(2022{\natexlab{a}})]{lu2022bayesian}
Jun Lu.
\newblock Bayesian low-rank interpolative decomposition for complex datasets.
\newblock \emph{arXiv preprint arXiv:2205.14825, Studies in Engineering and
  Technology}, 9\penalty0 (1):\penalty0 1--12, 2022{\natexlab{a}}.

\bibitem[Lu(2022{\natexlab{b}})]{lu2022comparative}
Jun Lu.
\newblock Comparative study of inference methods for interpolative
  decomposition.
\newblock \emph{arXiv preprint arXiv:2206.14542}, 2022{\natexlab{b}}.

\bibitem[Lu(2022{\natexlab{c}})]{lu2022matrix}
Jun Lu.
\newblock Matrix decomposition and applications.
\newblock \emph{arXiv preprint arXiv:2201.00145, Eliva Press},
  2022{\natexlab{c}}.

\bibitem[Lu \& Chai(2022)Lu and Chai]{lu2022robust}
Jun Lu and Christine~P Chai.
\newblock Robust {B}ayesian nonnegative matrix factorization with implicit
  regularizers.
\newblock \emph{arXiv preprint arXiv:2208.10053}, 2022.

\bibitem[Lu \& Ye(2022)Lu and Ye]{lu2022flexible}
Jun Lu and Xuanyu Ye.
\newblock Flexible and hierarchical prior for {B}ayesian nonnegative matrix
  factorization.
\newblock \emph{arXiv preprint arXiv:2205.11025}, 2022.

\bibitem[Marlin(2003)]{marlin2003modeling}
Benjamin~M Marlin.
\newblock Modeling user rating profiles for collaborative filtering.
\newblock \emph{Advances in neural information processing systems}, 16, 2003.

\bibitem[Martinsson et~al.(2011)Martinsson, Rokhlin, and
  Tygert]{martinsson2011randomized}
Per-Gunnar Martinsson, Vladimir Rokhlin, and Mark Tygert.
\newblock A randomized algorithm for the decomposition of matrices.
\newblock \emph{Applied and Computational Harmonic Analysis}, 30\penalty0
  (1):\penalty0 47--68, 2011.

\bibitem[Mnih \& Salakhutdinov(2007)Mnih and
  Salakhutdinov]{mnih2007probabilistic}
Andriy Mnih and Russ~R Salakhutdinov.
\newblock Probabilistic matrix factorization.
\newblock \emph{Advances in neural information processing systems}, 20, 2007.

\bibitem[Tulchinsky(2019)]{tulchinsky2019finding}
Igor Tulchinsky.
\newblock \emph{Finding Alphas: A quantitative approach to building trading
  strategies}.
\newblock John Wiley \& Sons, 2019.

\bibitem[Wang et~al.(2013)Wang, Wang, and Gao]{wang2013non}
Jim Jing-Yan Wang, Xiaolei Wang, and Xin Gao.
\newblock Non-negative matrix factorization by maximizing correntropy for
  cancer clustering.
\newblock \emph{BMC bioinformatics}, 14\penalty0 (1):\penalty0 1--11, 2013.

\end{thebibliography}
